\documentclass[final]{cvpr}

\usepackage{times}
\usepackage{epsfig}
\usepackage{graphicx}
\usepackage{amsmath}
\usepackage{amssymb}

\usepackage{bm}
\usepackage{booktabs}
\usepackage{siunitx}
\usepackage{stfloats}
\usepackage{multirow}
\usepackage{subcaption}

\usepackage[pagebackref=true,breaklinks=true,colorlinks,bookmarks=false]{hyperref}

\def\cvprPaperID{564} 
\def\confYear{CVPR 2021}

\begin{document}

\title{Dynamic Transfer for Multi-Source Domain Adaptation}

\author{Yunsheng Li\textsuperscript{1}, Lu Yuan\textsuperscript{2}, Yinpeng Chen\textsuperscript{2}, Pei Wang\textsuperscript{1}, Nuno Vasconcelos\textsuperscript{1}\\
\textsuperscript{1} UC San Diego, \textsuperscript{2} Microsoft\\
{\tt\small \{yul554,pew062,nvasconcelos\}@ucsd.edu, \{luyuan,yiche\}@microsoft.com}
}

\maketitle

\begin{abstract}
   
   Recent works of multi-source domain adaptation focus on learning a domain-agnostic model, of which the parameters are \textbf{static}. However, such a static model is difficult to handle conflicts across multiple domains, and suffers from a performance degradation in both source domains and target domain. In this paper, we present \textbf{dynamic transfer} to address domain conflicts, where the model parameters are adapted to samples. The key insight is that adapting model across domains is achieved via adapting model across samples. Thus, it breaks down source domain barriers and turns multi-source domains into a single-source domain. This also simplifies the alignment between source and target domains, as it only requires the target domain to be aligned with any part of the union of source domains. Furthermore, we find dynamic transfer can be simply modeled by aggregating residual matrices and a static convolution matrix. Experimental results show that, without using domain labels, our dynamic transfer outperforms the state-of-the-art method by more than $3\%$ on the large multi-source domain adaptation datasets -- DomainNet. Source code is at \href{https://github.com/liyunsheng13/DRT}{https://github.com/liyunsheng13/DRT}.
\end{abstract}

\section{Introduction}
\begin{figure}[t]
\centering
\includegraphics[width=0.95\linewidth]{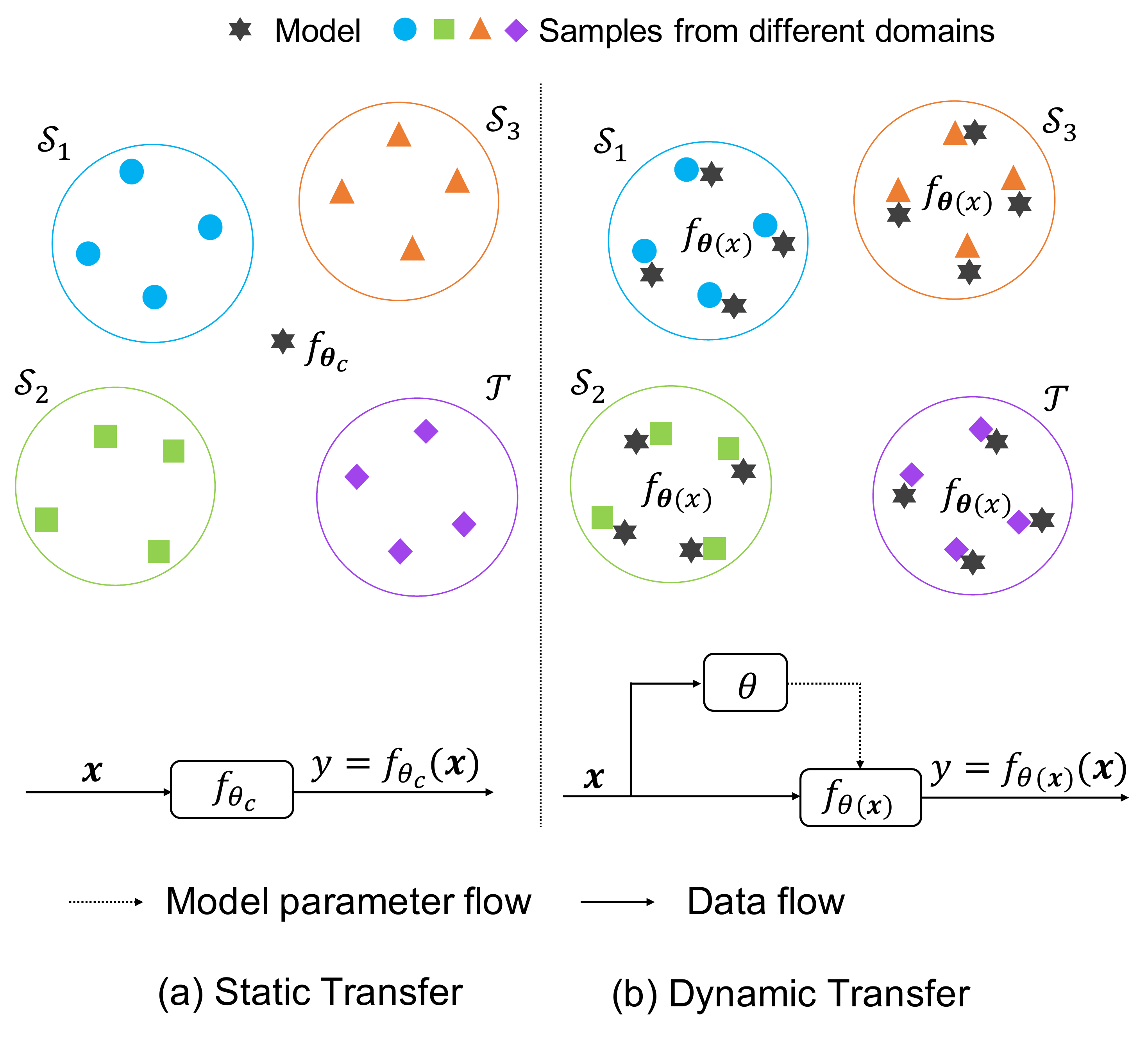}
\vspace{-0.5em}
\caption{\textbf{Static Transfer vs. Dynamic Transfer}. (a) `Static Transfer' implements domain adaptation with a static model $f_{\bm{\theta}_c}$, which has fixed parameters $\bm{\theta}_c$ to average domain conflict. (b) `Dynamic Transfer' ($f_{\bm{\theta}(\bm{x})}$) adapts the model parameters $\bm{\theta}(\bm{x})$ according to samples, which generates a different model per sample and turns multi-source domain adaptation into single-source domain adaptation. }
\label{fig:comb_teaser}
\vspace{-1.5em}
\end{figure}

\begin{figure}[t]
\centering
\includegraphics[width=0.98\linewidth]{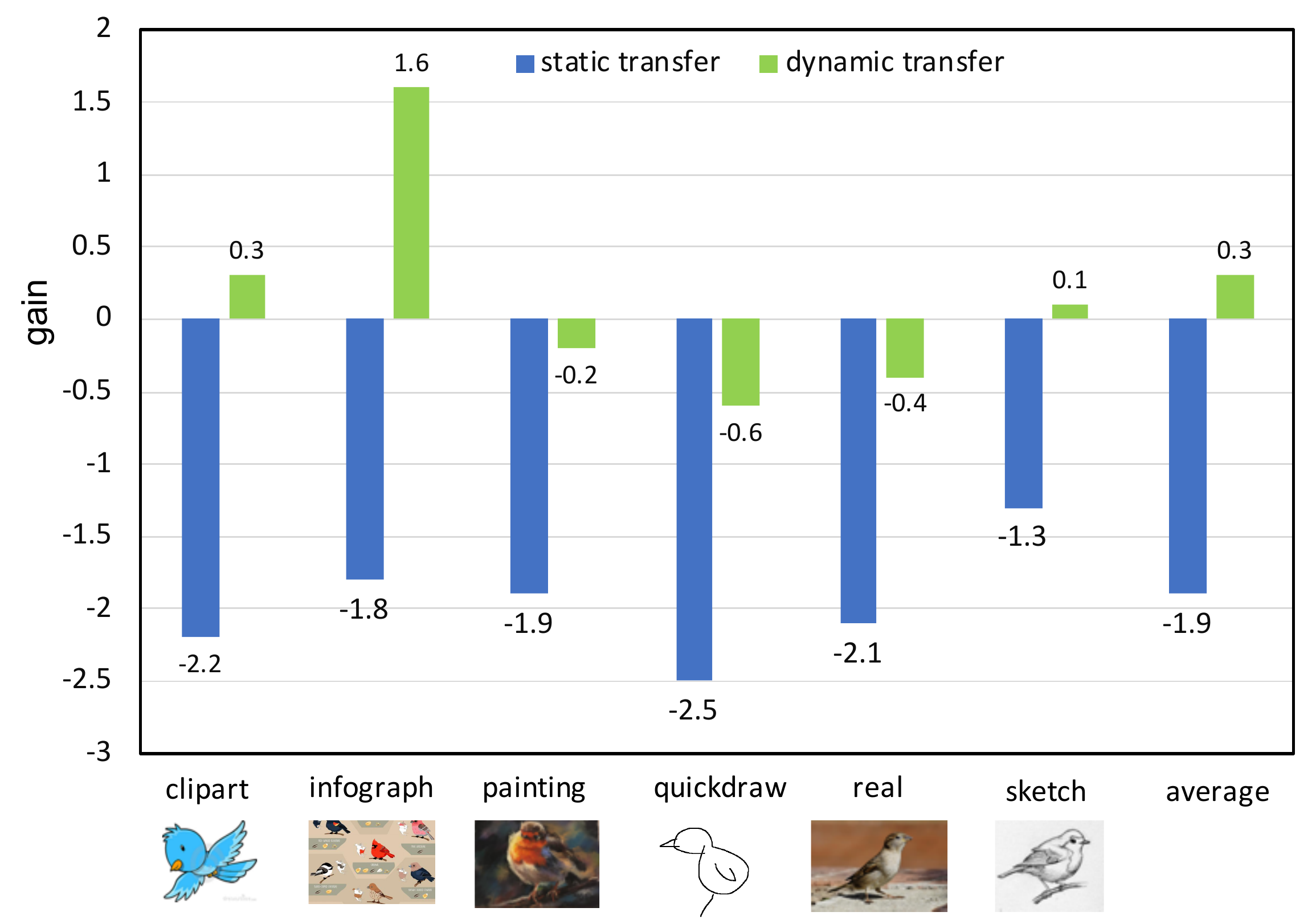}
\caption{\textbf{Static Transfer vs. Dynamic Transfer on the performance degradation of source domains} compared to the oracle results. Both transfer models are tested across source domains.
A clear performance degradation ($1.9\%$ in average) exists when using static transfer, indicating the conflicts across domains. The degradation is significantly reduced when using dynamic transfer, as it handles the domain variations well. (Best view in color)}
\label{fig:hist_teaser}
\vspace{-1.0em}
\end{figure}

Multi-source domain adaptation addresses the adaptation from multiple source domains to a target domain. It is challenging because a clear domain discrepancy exists not only between source and target domains, but also among multiple source domains (see exemplar images in Figure \ref{fig:hist_teaser}). This suggests that successful adaptation requires significant {\it elasticity\/} of the model to adapt. 
A nature way to achieve this elasticity is to make model dynamic i.e. the mapping implemented by the model should vary with the input sample.

This hypothesis has not been explored by existing work, e.g. \cite{peng2019moment,wang2020learning}, which instead aims to learn a domain agnostic model $f_{\bm{\theta}_c}$, of {\it static} parameters $\bm{\theta}_c$, that works well for all source  $\{\mathcal{S}_1,\mathcal{S}_2,...,\mathcal{S}_N\}$ and target  $\mathcal{T}$ domains. We refer to this approach as {\it static transfer.\/} As illustrated in Figure \ref{fig:comb_teaser} (a), the model implements a fixed mapping across all domains. However, learning a domain agnostic model is difficult, since different domains can give rise to very different image distributions. When forcing a model to be domain agnostic, it essentially averages the domain conflict. Thus the performance drops on each source domain. This is validated by our preliminary study. As shown in Figure \ref{fig:hist_teaser}, compared to the optimal model per domain, the static transfer model consistently degrades in each source domain.

In this paper, we propose {\it{dynamic transfer}} to address this issue. As shown in Figure~\ref{fig:comb_teaser}(b), it contains a parameter predictor that changes the model parameters on a per-sample basis, i.e. implements mapping $f_{\bm{\theta}(\bm{x})}$. It has the advantage of not requiring the definition of domains or the collection of domain labels. In fact, it unifies the problems of single-source and multi-source domain adaptation. By breaking down source domain barriers, it turns multiple source domain adaptation into a single-source domain problem. The only difference is the complexity of this domain. 

The key insight is that adapting model according to domains is achieved statistically by adapting model per sample, since each domain is viewed as a distribution of image samples. The dynamic transfer learns how to adapt the model's parameters and fit to the union of source domains. Thus the alignment between source domains and target domain is significantly simplified, as it is no longer necessary to pull all source domains together with the target domain. In this case, as long as the target domain is aligned to any part of the source domains, the model can be easily adapted to the target samples.

When compared to the domain adaption literature, dynamic transfer introduces a significant paradigm shift. In the literature, most works assume the static network of Figure~\ref{fig:comb_teaser}(a) and focus on loss functions. The goal is to define losses that somehow ``pull all the domains" together into a shared latent representation. The problem is that the domains are usually very different at the network input. Hence, the force introduced by the loss at the output, to bring them together, is counter-balanced by an input force to keep them apart. This usually leads to a difficult optimization and compromises adaptation performance. The introduction of a dynamic network, as in Figure~\ref{fig:comb_teaser}(b), enables a more elastic mapping. In this case, it is not necessary to pull all domains together. The model adaptation given by the dynamic transfer can be generalized to target domain easily when the target domain is shifted to the space formed by entire source domain. In this way, dynamic transfer shifts the focus of the domain adaptation problem from the design of good {\it loss functions\/} to the design of good {\it network architectures\/} for dynamic transfer.

An immediate difficulty is that the architecture of Figure~\ref{fig:comb_teaser}(b) can be very hard to train, since the parameter predictor cannot generate all parameters for a large model.
The question is whether it is possible to perform the model adaptation by only modifying a small subset of parameters on a sample basis. In this work, we show that this is indeed possible by the addition of dynamic residuals to the convolution kernels of a static network.
Since the residual blocks can be much smaller than the static ones, this has both very low additional computational cost (less than 0.1\%) to aggregate dynamic residuals with static kernel and little tendency to overfit. However, it is shown to significantly enhance the domain adaptation performance. 
Experimental results show that the proposed {\it dynamic residual transfer\/} (DRT) can model domain variation in source domains (see Figure \ref{fig:hist_teaser}) and outperform its static counterpart (MCD \cite{saito2018maximum} method) by a large margin (11.2\% on DomainNet). Compared to state-of-the-art multi-source domain adaptation methods \cite{wang2020learning}, it achieves a sizeable gain ($3.9\%$) with a much simpler loss function and training algorithm.

\section{Related Work}
\noindent{\bf{Single-Source Domain Adaptation}} approaches adapt a model from a source to a target domain. A common method is to minimize the distance between the two domains. While some methods \cite{cariucci2017autodial,sun2016deep} minimize distance functions defined in terms of first and second order data statistics, others learn a latent space shared across domains by adversarial learning \cite{tzeng2017adversarial,saito2018maximum,hoffman2018cycada,li2019bidirectional}. 
Although these methods are effective for single-source domain distributions and relatively simple datasets (such as VisDA \cite{peng2017visda} or Office-31 \cite{saenko2010adapting}), they are not competitive for the multi-source domain adaptation problem, due to a more  complex data distribution.

\noindent{\bf{Multi-Source Domain Adaptation}} considers the domain adaptation problem when the source contains domains with a variety of styles. \cite{yang2007cross} pioneered this problem by adaptively picking the best among a set of hypothesis learned for different source domains. \cite{blitzer2008learning} derived an upper bound on the classification error achievable in the target domain, based on the $\mathcal{H}\rm \Delta \mathcal{H}$ divergence. Several methods have been proposed after the introduction of deep learning. Some of these align domains pair-wise. \cite{xu2018deep} uses a discriminator to align each source domain with the target domain, while \cite{peng2019moment} matches moments across all pairs of source and target domains. These methods learn one classifier per domain and use their weighted combination to predict the class of target samples. \cite{li2020mutual} uses mutual learning techniques to align feature distributions among pairs of source and target domains. 
Other methods focus on the joint alignment of the feature distributions of all domains. \cite{wang2020learning} models interactions between domains with a knowledge graph. Target sample predictions are based on both their features and relationship to different domains. \cite{li2020online} proposes a meta-learning technique to search the best initial conditions for multi-source domain adaptation. \cite{yang2020curriculum} uses an auxiliary network to predict the transferability of each source sample and use it as a weight to learn a domain discriminator. All these works use a static transfer model. In this paper, we propose that the model should instead be dynamic, i.e. a function that changes with samples, and show that this can significantly enhance multi-source domain adaptation.

\noindent{\bf{Dynamic Networks}} have architectures based on blocks \cite{lin2017runtime,wu2018blockdrop,yang2019condconv,chen2019dynamic} or channels \cite{hu2018squeeze,cao2019gcnet,wang2019towards} that change depending on the input sample. \cite{lin2017runtime,wu2018blockdrop} proposed an input dependent block path that decides whether a network block should be kept or dropped. \cite{yang2019condconv,chen2019dynamic} widen the network by adding new parallel blocks and train an attention module to choose the best combination of features dynamically. \cite{hu2018squeeze,cao2019gcnet,wang2019towards} rely on feature based attention modules that reweigh features depending on the input example. \cite{wang2019towards} shows that, for object detection, objects from different domains are best detected with domain dependent features. In this paper, we propose a dynamic convolution residual branch, which adds an input-dependent residual matrix to a static kernel, to implement dynamic multi-source domain adaptation.

\section{Method}

In this section, we introduce dynamic transfer for multi-source domain adaptation, in which the model is adaptive to the domain implicitly, but adaptive to the input explicitly. It not only has better performance, but also turns multi-source domains into a single-source domain.

\subsection{Multi-Source Domain Adaptation}

Multi-source domain adaptation (MSDA) aims to transfer a model learned on a source data distribution drawn from several domains $\mathcal{S} = \{\mathcal{S}_1,...,\mathcal{S}_N\}$ to a target domain $\mathcal{T}$. While the following ideas can be applied to various tasks, we consider a classification model
$f_{\bm{\theta}}$, of parameters $\bm{\theta}$, which maps images $\bm{x} \in {\cal X}$ to class predictions $y \in {\cal Y} = \{1, \ldots, C\}$, where $C$ is the number of classes and $\cal X$ is some image space. The goal is to adapt the parameters $\bm{\theta}$ of a model learned from a dataset ${\cal D}^{\cal S} = \{(\bm{x}^{\cal S}_i,{\bf y}_i)\}_{i=1}^{{N}_{\cal S}}$ of examples from the source distribution $\cal S$ (${\bf y}_i$ is the one-hot encoding of the label of example $\bm{x}^{\cal S}_i$) to a dataset ${\cal D}^T = \{\bm{x}^{\cal T}_i\}_{i=1}^{{N}_{\cal T}}$ of unlabeled examples from the target distribution. Note that, in the most general formulation of the problem, the domain of origin of each source example, $(\bm{x}^{\cal S}_i,{\bf y}_i)$ is {\it unknown\/}. This is ignored by many approaches e.g. \cite{peng2019moment,li2020mutual}, that assume a source dataset ${\cal D}^{\cal S} = \{(\bm{x}^{\cal S}_i,{\bf y}_i, z_i)\}_{i=1}^{{N}_{\cal S}}$ contains domain labels $z_i \in \{1, \ldots, N\}$ and aligning pairs of domains. We refer to this a domain supervised multi-source domain adaptation.

\subsection{Static vs. Dynamic Transfer}
The model $f_{\bm{\theta}}$ is denoted static or dynamic depending on whether the model parameters $\bm{\theta}$ vary with samples $\bm{x}$. Static models have constant parameters $\bm{\theta}=\bm{\theta}_c$, while dynamic models have parameters $\bm{\theta}=\bm{\theta}(\bm{x})$ that depend on $\bm{x}$. In the case of deep networks, this implies that layer transfer functions depend on the input $\bm{x}$. Figure~\ref{fig:comb_teaser} illustrates the static transfer and dynamic transfer model built for multi-source domain adaptation. 

\noindent{\bf Static Transfer.} Static transfer, shown on Figure \ref{fig:comb_teaser}(a), consists of learning of a single model $f_{\bm{\theta}_c}$ that is applied to {\it all\/} examples from source and target domains. The model might, for instance, map images into a latent space where all the distributions are aligned. Since the big variation among the input samples, this is a difficult problem and the model $f_{\bm{\theta}_c}$ usually has sub-optimal performance on all domains.

\noindent \textbf{Dynamic Transfer.} In this case the model parameters are a function of the input example $\bm{x}$ directly, i.e. the model has the form $f_{\bm{\theta}(\bm{x})}$ where $\bm{x} \in \mathcal{S}_1\bigcup\dots\bigcup\mathcal{S}_N\bigcup\mathcal{T}$. This is illustrated in Figure~\ref{fig:comb_teaser}(b), where there exists a model per sample. Compared to the static transfer, dynamic transfer varies the model according to sample explicitly and chooses domains {\it implicitly,\/} relying on the distribution of samples $\bm{x}$. Dynamic transfer learns to adapt the parameters to fit the model to the distribution formed by the union of source domains. The target domain is not required to be aligned with any specific domains ${\cal S}_i$ and there are no rigid domain boundaries. The model parameters $\bm{\theta}(\bm{x})$ can be similar for examples from different domains and different for examples from the same domain. 

The key insight is that adapting model per domain is achieved statistically by adapting model per sample, as each domain can be considered as a distribution of image samples. The dynamic transfer learns to adapt model parameters over samples in the union of all source domains. This simplifies the alignment between source and target domains, as it is not necessary to pull all source domains and target domain together. As long as the target domain is aligned with any part of the union of source domains, the model can be easily adapted to the target samples. 

Dynamic transfer has two advantages over static transfer. First, it turns multi-source domains into a single-source domain, voiding the need for domain labels. Second, it simplifies learning, since domain labels can be arbitrary. In practice, any ``domain" can contain a mixture of unlabeled sub-domains and some of these can be shared by multiple ``domains". Due to this, explicit assignment of data to domains can be difficult, and models learned over single domain can loose access to shared sub-domain data. 

\subsection{Dynamic Residual Transfer}
The main difficulty of dynamic transfer is the model $f_{\bm{\theta}(\bm{x})}$ can be difficult to learn. Given the large number of parameters of modern networks, it is impossible to simply predict all parameter values at inference time. The key is to restrict the model's dependence on input $\bm{x}$ to a {\it small number of parameters\/}. To guarantee this, we propose a model composed by a {\it static network\/} and {\it dynamic residual blocks\/}
\begin{equation}
    f_{\bm{\theta}(\bm{x})} = f_{0} + \Delta f_{\bm{\theta}(\bm{x})},
    \label{equ:dynamic_res_model}
\end{equation}
where $f_{0}$ represents the static component and $\Delta f_{\bm{\theta}(\bm{x})}$ the dynamic residual that depends on the input sample $\bm{x}$. As usual, the residual is implemented by adding residual blocks to the various network layers. Since the static component $f_0$ is shared by all samples, 
static transfer is a special case of the proposed approach, where $\Delta f_{\bm{\theta}(\bm{x})}=0$. This approach is denoted as {\it dynamic residual transfer\/}  (DRT).

To implement DRT in convolution neural networks (CNNs), we represent a $k \times k$ convolution kernel as a $C_{out} \times C_{in}k^2$ weight matrix, where $C_{in}$ and $C_{out}$ are the number of input and output channels. We ignore bias terms in this discussion for the sake of brevity.
DRT is implemented by applying Equation \ref{equ:dynamic_res_model} to each convolution kernel in a CNN, i.e. defining the network convolutions as
\begin{equation}
    \bm{W}(\bm{x}) = \bm{W}_0 + \Delta \bm{W}(\bm{x}),
\end{equation}
where $\bm{W}_0$ is a static convolution kernel matrix, and $\Delta \bm{W}(\bm{x})$ a dynamic residual matrix. We next discuss several possibilities for the latter.

\noindent\textbf{Channel Attention:} in this case, the residual only rescales the output channels of $\bm{W}_0$. This is implemented as
\begin{align}
    \Delta \bm{W}(\bm{x}) &= \bm{\Lambda}(\bm{x})\bm{W}_0,
    \label{equ:dy_scale}
\end{align}
where $\bm{\Lambda}(\bm{x})$ is a diagonal $C_{out} \times C_{out}$ matrix, whose entries are functions of $\bm{x}$. This can be seen as a dynamic feature-based attention mechanism.

\noindent \textbf{Subspace Routing:} the dynamic residual is a linear combination of $K$ static  matrices $\bm{\Phi}_i$ 
\begin{align}
    \Delta \bm{W}(\bm{x}) &=  \sum_{i=1}^K\pi_i(\bm{x})\bm{\Phi}_i,
    \label{equ:dy_res}
\end{align}
whose weights depend on $\bm{x}$. The matrices $\bm{\Phi}_i$ can be seen as a basis for CNN weight space, although they are not necessarily linearly independent. And the dynamic coefficients $\pi_i(\bm{x})$ can be seen as the projections of the residual matrix in the corresponding weight subspaces. By choosing these projections in an input dependent manner, the network chooses different feature subspaces to route different $\bm{x}$. 

\begin{figure}[t]
\centering
\includegraphics[width=0.98\linewidth]{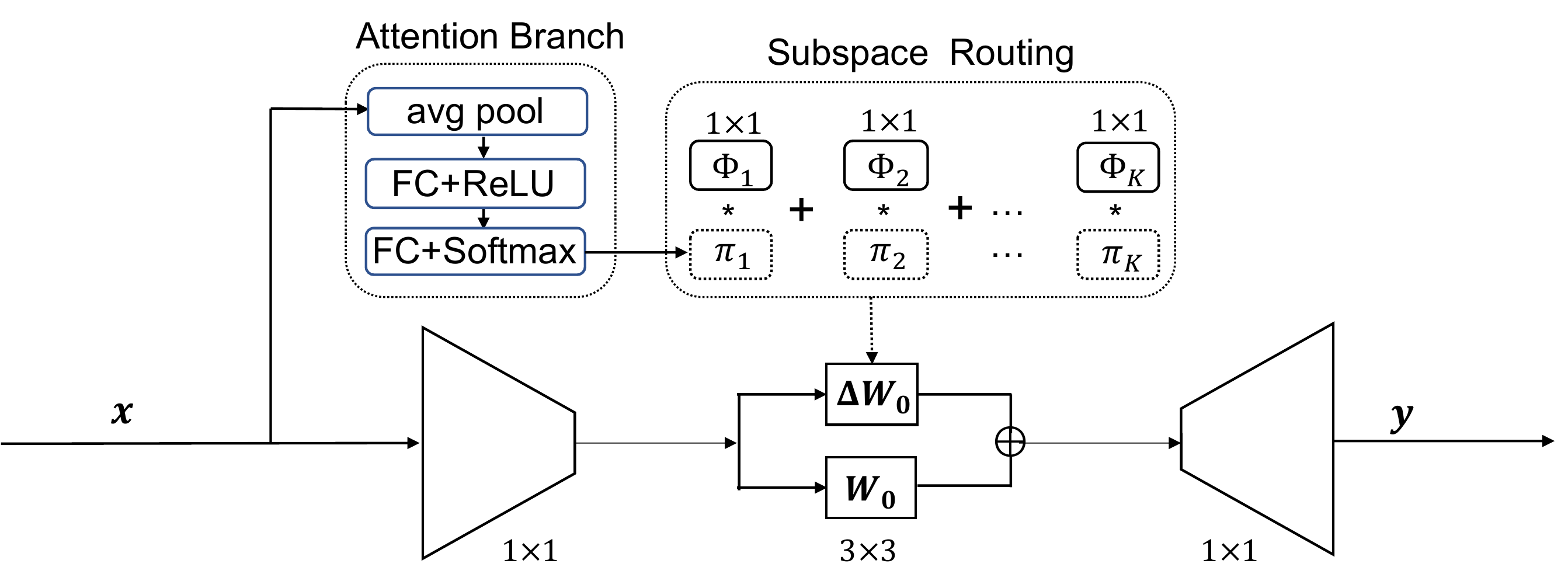}
\caption{Subspace routing of DRT: dynamic coefficients are generated by a dynamic branch given the input $\bm{x}$. Each dynamic coefficient $\pi_i(\bm{x})$ is then multiplied by a matrix $\bm{\Phi}_i$, and the $K$ matrices are aggregated as the residual kernel- $\bm{\Delta W}_0(\bm{x})$. For channel attention, softmax is replaced by sigmoid and the resulting coefficients in $\bm{\Lambda}(\bm{x})$ are multiplied to corresponding channels of $\bm{W}_0$.}
\label{fig:dynamic_block}
\vspace{-1.0em}
\end{figure}
To reduce the number of parameters and computation, the matrices can be further simplified into $1 \times 1$ convolution kernels and applied to the narrowest layer of the bottleneck architecture in ResNet \cite{he2016deep}. In this case, only $C_{in}$ rows of $\bm{\Phi}_i$ are non-zero. 

\noindent \textbf{Combination:} the two mechanisms are combined into
\begin{align}
    \Delta \bm{W}(\bm{x}) &=  \bm{\Lambda}(\bm{x})\bm{W}_0 + \sum_{i=1}^K\pi_i(\bm{x})\bm{\Phi}_i.
    \label{equ:dy_aff}
\end{align}
Similar to squeeze-and-excitation block \cite{squeezenet16}, the dynamic coefficients $\bm{\Lambda}(\bm{x})$ and $\{\pi_i(\bm{x})\}$ are implemented by a light-weight attention branch that includes average pooling and two fully connected layers (See Figure \ref{fig:dynamic_block}). A sigmoid is used to normalize $\bm{\Lambda}(\bm{x})$ and a softmax to normalize $\{\pi_i(\bm{x})\}$. As explained by \cite{chen2019dynamic,yang2019condconv}, the extra FLOPs caused by dynamic coefficient generation and residual aggregation of dynamic transfer is negligible (less than 0.1\% in our implementation) compared to the static model.

\subsection{Learning}
As usual for domain adaptation problems, the DRT network is learned with a combination of two losses,
\begin{equation}
    \mathcal{L} = \mathcal{L}_{ce} + \lambda \mathcal{L}_{d},
    \label{equ:loss}
\end{equation}
where $\lambda$ is a hyperparameter that controls the trade-off between the loss components. The first loss
\begin{align}
    \mathcal{L}_{ce} = \frac{1}{N_\mathcal{S}}\sum_{i=1}^{N_\mathcal{S}} 
    \mathbf{y}_i^T \log f_{\bm{\theta}(\bm{x}_i^\mathcal{S})}(\bm{x}_i^\mathcal{S}),
    \label{equ:ce_loss}
\end{align}
is the cross entropy loss over the source data ${\cal D}^{\cal S}$. The second is a domain alignment loss that encourages the minimization of the distance between source and target domains
\begin{align}
    \mathcal{L}_{d}= {\cal H} \left(f_{\bm{\theta}({\cal D}^{\cal S})}({\cal D}^{\cal S}), f_{\bm{\theta}({\cal D}^{\cal T})}({\cal D}^{\cal T})\right),
    \label{equ:da_loss}
\end{align}
where ${\cal D}^{\cal T}$ is the target data and $\cal H$ a measure of discrepancy between feature distributions of the source and target domains. $\cal H$ can be any distance function previously proposed for domain adaptation, e.g. the MMD \cite{long2015learning} or adversarial learning \cite{tzeng2017adversarial}. Note that the two losses above operate on the entire source dataset ${\cal D}^{\cal S}$, i.e. there is no need for domain labels and not even a difference between the single domain and multiple domains adaptation problems.
For the domain alignment losses commonly used in multi-source domain adaptation, Equation \ref{equ:da_loss} also does require the evaluation of pairwise distances between all source domains and target domain, which is not necessary in dynamic transfer. 

\section{Experiments}
\label{sec:exp}
In this section, adaptation performance of DRT is evaluated.

\begin{table*}[t]
	\centering
	\small
	\setlength{\tabcolsep}{2.5pt}
	\begin{tabular}{cccccccc}
		\hline
		Models & \begin{tabular}{cc}inf,pnt,qdr \\ rel,skt $\rightarrow$ clp\end{tabular}& \begin{tabular}{cc}clp,pnt,qdr \\ rel,skt $\rightarrow$ inf\end{tabular} & \begin{tabular}{cc}clp,inf,qdr \\ rel,skt $\rightarrow$ pnt\end{tabular} & \begin{tabular}{cc}clp,inf,pnt \\ rel,skt $\rightarrow$ qdr\end{tabular} & \begin{tabular}{cc}clp,inf,pnt \\ qdr,skt $\rightarrow$ rel\end{tabular} & \begin{tabular}{cc}clp,inf,pnt \\ qdr,rel $\rightarrow$ skt\end{tabular} & Avg\\
		
		\specialrule{.1em}{.05em}{.05em}
		static & 54.3$\pm$0.64& 22.1$\pm$0.70 & 45.7$\pm$0.63 & 7.6$\pm$0.49 & 58.4$\pm$0.65 & 43.5$\pm$0.57 & 38.5$\pm$0.61\\
        Channel Attention & 67.8$\pm$0.46 & 30.9$\pm$0.85 & 57.1$\pm$0.36 & 6.9$\pm$1.12 & 66.7$\pm$0.42 & 57.4$\pm$0.33 & 47.8$\pm$0.59\\
		
		Subspace Routing & \textbf{69.7}$\pm$0.24& 31.0$\pm$0.56 & \textbf{59.5}$\pm$0.43 & 9.9$\pm$1.03 & \textbf{68.4}$\pm$0.28 & \textbf{59.4}$\pm$0.21 & \textbf{49.7}$\pm$0.46\\
		Combination & 69.1$\pm$0.35 & \textbf{31.6}$\pm$0.61 & 58.2$\pm$0.25 & \textbf{11.9}$\pm$0.96 & 67.8$\pm$0.36 & 58.8$\pm$0.44 & 49.6$\pm$0.50\\
		\specialrule{.1em}{.05em}{.05em}
	\end{tabular}
	\caption{Comparison of \textbf{different implementations for dynamic residual transfer}: Channel Attention (Equation \ref{equ:dy_scale}), Subspace Routing (Equation \ref{equ:dy_res}) and Combination (Equation \ref{equ:dy_aff}).}
	\label{tab:dy_abl}
	\vspace{-1.0em}
\end{table*}

\vspace{-0.5em}
\subsection{Datasets and Experimental Settings:}
Following \cite{peng2019moment}, we consider two datasets, Digit-five and DomainNet \cite{peng2019moment}, which contain images from several domains but shared classes. Each domain is alternatively used as the target domain and the remaining ones as the source domain. All experiments are repeated with $5$ times and mean and variance are reported.

\vspace{0.5mm}
\noindent \textbf{Digit-five:} Digit-five contains digit images from $5$ domains: MNIST \cite{lecun1998gradient} (mt), Synthetic \cite{ganin2014unsupervised} (sy), MNIST-M \cite{ganin2014unsupervised} (mm), SVHN \cite{netzer2011reading} (sv) and USPS \cite{ganin2014unsupervised} (up). These domains contribute $25,000$ images for training and $9000$ for validation, with the exception of USPS which uses $29752$ and $1860$, respectively. Since these datasets are relatively small, LeNet \cite{lecun1998gradient} is used as the backbone model. A dynamic residual is added on each convolutional layer. The model is trained from scratch with initial learning rate $0.002$ and SGD optimizer. The learning rate is decayed by $0.1$ every $100$ epochs and decreased to $2e-5$ in $300$ epochs. 

\vspace{0.5mm}
\noindent \textbf{DomainNet:} DomainNet \cite{peng2019moment} is a dataset with $0.6$ million images of $345$ classes from $6$ domains of different image styles: clipart (clp), infograph (inf), painting (pnt), quickdraw (qdr), real (rel) and sketch (skt). Results are obtained with ImageNet \cite{deng2009imagenet} pretrained ResNet-101 \cite{he2016deep}. The dynamic residual is only added on the $3\times 3$ kernel of each bottleneck block. The networks are trained with SGD for $15$ epochs with initial learning rate of $0.001$ and batch size as $64$. The learning rate is decayed by $0.1$ every $5$ epochs. 

\subsection{Ablation Study}
An ablation study was performed on DomainNet to evaluate the three key components of DRT: (a) the three implementations of the dynamic transfer, (b) the number of basis used for subspace routing (Equation \ref{equ:dy_res}), and (c) different alignment losses ${\cal L}_d$. The default model uses subspace routing with $K=4$ and is trained with the MCD \cite{saito2018maximum} loss. 

\vspace{0.3em}
\noindent \textbf{DRT Implementations: } 
Table \ref{tab:dy_abl} shows that all implementations of DRT 
have significantly better adaptation performance than the static model. The average gains are of $9.3\%$ for channel attention, $10.8\%$ for combined and $11.2\%$ for subspace routing. The weaker performance of channel attention suggests that it is not enough to re-scale the features of the static model. Routing the input $\bm{x}$ through different subspaces appears to be more effective, although the differences are not staggering. While combining the two approaches has no additional overall benefit, the combination was beneficial for specific transfer problems. 
When `infograph' and `quickdraw' were used as target domains, the combination model outperformed subspace routing. Since these are the hardest transfer problems, this suggests that the enhanced dynamics of the combined implementation can be beneficial as the domain gap increases. It is because the enhanced dynamics make the model more elastic. Therefore, it is more likely to adapt models to target domain with larger gap. On the other hand, for the problems of smaller domain gap, stronger dynamics can cause the model to overfit to the source domain, as is the case for the remaining target domains. More experiments on datasets with more domains will likely be needed to resolve this question. In any case, subspace routing and the combination model have similar performance.

\vspace{0.3em}
\noindent \textbf{Number of Residual Basis.} The impact of the number of basis $K$ used in Equation \ref{equ:dy_res} for subspace routing is ablated. For different values of $K\in\{2,4,6,8\}$, DRT achieves $\{48.8,49.7,49.5,49.3\}$, all of which improve the adaptation performance of static transfer ($38.5\%$) by a large margin (more than 10\%). Best performance is achieved with $K=4$, although the results are not highly sensitive to this parameter. 

\vspace{0.3em}
\noindent \textbf{Alignment Loss Function.} Three different domain alignment losses with different forms of $\cal H$ (see Equation \ref{equ:da_loss}) were compared: ADDA \cite{tzeng2017adversarial}, MCD \cite{saito2018maximum} and $M^3SDA$ \cite{peng2019moment}. They are representative of previously proposed losses for reducing single-source domain shift at the domain level and class level, and multi-source domain shift, respectively.

\begin{table*}[t]
	\centering
	\small
	\setlength{\tabcolsep}{2.5pt}
	\begin{tabular}{l|ccccccc}
		\specialrule{.1em}{.05em}{.05em}
		 ${\cal L}_d$ & \begin{tabular}{cc}inf,pnt,qdr \\ rel,skt $\rightarrow$ clp\end{tabular}& \begin{tabular}{cc}clp,pnt,qdr \\ rel,skt $\rightarrow$ inf\end{tabular} & \begin{tabular}{cc}clp,inf,qdr \\ rel,skt $\rightarrow$ pnt\end{tabular} & \begin{tabular}{cc}clp,inf,pnt \\ rel,skt $\rightarrow$ qdr\end{tabular} & \begin{tabular}{cc}clp,inf,pnt \\ qdr,skt $\rightarrow$ rel\end{tabular} & \begin{tabular}{cc}clp,inf,pnt \\ qdr,rel $\rightarrow$ skt\end{tabular} & Avg\\
		
		\hline
		 Source Only & 52.1$\pm$0.51 & 23.4$\pm$0.28 & 47.7$\pm$0.96 & \textbf{13.0}$\pm$0.72 & 60.7$\pm$0.23 & 46.5$\pm$0.56 & 40.6$\pm$0.56\\
		Source Only + \textbf{DRT} & \textbf{63.1}$\pm$0.62 & \textbf{25.9}$\pm$0.84 & \textbf{48.4}$\pm$1.02 & 6.4$\pm$0.98 & \textbf{66.4}$\pm$0.54 & \textbf{46.8}$\pm$0.44 & \textbf{42.8}$\pm$0.74\\
		\hline
		ADDA \cite{tzeng2017adversarial} & 47.5$\pm$0.76 & 11.4$\pm$0.67 & 36.7$\pm$0.53 & \textbf{14.7}$\pm$0.50 & 49.1$\pm$0.82 & 33.5$\pm$0.49 & 32.2$\pm$0.63\\
		ADDA+\textbf{DRT}& \textbf{63.6}$\pm$0.52 & \textbf{27.6}$\pm$0.43& \textbf{52.3}$\pm$0.68 & 8.2$\pm$1.44 & \textbf{67.9}$\pm$0.42 & \textbf{49.6}$\pm$0.33 & \textbf{44.9}$\pm$0.64\\
		\hline
		MCD \cite{saito2018maximum}& 54.3$\pm$0.64& 22.1$\pm$0.70 & 45.7$\pm$0.63 & 7.6$\pm$0.49 & 58.4$\pm$0.65 & 43.5$\pm$0.57 & 38.5$\pm$0.61\\
		MCD+\textbf{DRT} & \textbf{69.7}$\pm$0.24& \textbf{31.0}$\pm$0.56 & \textbf{59.5}$\pm$0.43 & \textbf{9.9}$\pm$1.03 & \textbf{68.4}$\pm$0.28 & \textbf{59.4}$\pm$0.21 & \textbf{49.7}$\pm$0.46\\
		\specialrule{.1em}{.05em}{.05em}
		M$^3$SDA-$\beta$ \cite{peng2019moment} & 58.6$\pm$0.53 & 26.0$\pm$0.89 & 52.3$\pm$0.55 & 6.3$\pm$0.58 & 62.7$\pm$0.51 & 49.5$\pm$0.76 & 42.6$\pm$0.64\\
		M$^3$SDA-$\beta$+\textbf{DRT}& \textbf{67.4}$\pm$0.52 & \textbf{31.3}$\pm$0.83 & \textbf{56.5}$\pm$0.67 & \textbf{13.6}$\pm$0.34 & \textbf{66.9}$\pm$0.42 & \textbf{56.8}$\pm$0.49 & \textbf{48.8}$\pm$0.55\\
		\specialrule{.1em}{.05em}{.05em}
	\end{tabular}
	\caption{\textbf{Static transfer vs. Dynamic transfer} evaluated on DomainNet with different domain alignment loss functions.}
	\label{tab:loss_abl}
	\vspace{-0.5em}
\end{table*}

Table \ref{tab:loss_abl} shows that dynamic residual transfer (DRT) outperforms static transfer for all loss functions, by a large margin (12.7\%, 11.2\%, 6.2\% respectively). Its improved performance is in part, due to the fact that DRT takes a much larger advantage of the domain alignment losses. It confirms our claim that DRT simplifies the domain alignment by unifying all source domains into a single domain. Thus the target samples are more likely to be aligned with the union of source domains and the same alignment loss will give more benefits to the dynamic model than the static one.
However, the gains over the `source only' case, where no alignment loss ${\cal L}_d$ is used in Equation \ref{equ:loss}, is only $2\%$. It means alignment loss is very critical for dynamic transfer. Without alignment loss, even though the model can adapt to the entire source domain very well, it can hardly adapt to target samples due to a large domain gap.

These conclusions also apply to the individual transfer problems, except when `quickdraw' is the target domain. In this case,  DRT is only effective with the M$^3$SDA-$\beta$ \cite{peng2019moment} loss. 
It is because when `quickdraw' is the target domain, the domain discrepancy is much larger and makes it harder for DRT to adapt model to this domain. Thus, M$^3$SDA-$\beta$ \cite{peng2019moment} which proposed a more powerful alignment loss, can shift `quickdraw' closer to source domains and works better with DRT. However, the strong alignment loss will cause `over-alignment' for the domains e.g. `clipart' that have much smaller gap. The `over-alignment' reduces the adaptability of the dynamic model, which causes performance degradation compared to that given by simpler alignment losses e.g. MCD.

\subsection{Comparisons to the state-of-the-art}
DRT was compared to the results in the literature for Digit Five and DomainNet dataset. In these experiments, DRT is implemented with subspace routing ($4$ basis), using the MCD loss \cite{saito2018maximum}, and $\lambda=50$ in Equation \ref{equ:loss}.

\begin{table*}[t!]
	\centering
	\small
	\begin{tabular}{ccccccc}
		\specialrule{.1em}{.05em}{.05em}
		Models & \begin{tabular}{cc}mm,up,sv \\sy $\rightarrow$ mt\end{tabular} 
		       & \begin{tabular}{cc}mt,up,sv \\sy $\rightarrow$ mm\end{tabular}  
		       & \begin{tabular}{cc}mt,mm,sv \\sy $\rightarrow$ up\end{tabular}   
		       & \begin{tabular}{cc}mt,mm,up \\sy $\rightarrow$ sv\end{tabular} 
		       & \begin{tabular}{cc}mt,mm,up \\sv $\rightarrow$ sy\end{tabular} & Avg\\
		\hline
		Source Only & 63.37$\pm$0.74 & 90.50$\pm$0.83 & 88.71$\pm$0.89&63.54$\pm$0.93 &82.44$\pm$0.65&77.71$\pm$0.81\\
		\hline
		DANN \cite{ganin2015unsupervised} & 71.30$\pm$0.56&97.60$\pm$0.75&92.33$\pm$0.85&63.48$\pm$0.79&85.34$\pm$0.84&82.01$\pm$0.76\\
		ADDA \cite{tzeng2017adversarial} & 71.57$\pm$0.52&97.89$\pm$0.84&92.83$\pm$0.74&75.48$\pm$0.48&86.45$\pm$0.62&84.84$\pm$0.64\\
		MCD \cite{saito2018maximum} & 72.50$\pm$0.67&96.21$\pm$0.81&95.33$\pm$0.74&78.89$\pm$0.78&87.47$\pm$0.65&86.10$\pm$0.73\\
		DCTN \cite{xu2018deep} & 70.53$\pm$1.24& 96.23$\pm$0.82&92.81$\pm$0.27&77.61$\pm$0.41&86.77$\pm$0.78&84.79$\pm$0.72\\
		M$^3$SDA-$\beta$ \cite{peng2019moment} & 72.82$\pm$1.13&98.43$\pm$0.68&96.14$\pm$0.81&81.32$\pm$0.86&89.58$\pm$0.56&87.65$\pm$0.75\\
		CMSS \cite{yang2020curriculum} & 75.3$\pm$0.57 & 99.0$\pm$0.08 & 97.7$\pm$0.13 & \textbf{88.4}$\pm$0.54 & 93.7$\pm$0.21 & 90.8$\pm$0.31\\
		\hline
		\textbf{DRT} & \textbf{81.03}$\pm$0.34 &\textbf{99.31}$\pm$0.05 &\textbf{98.40}$\pm$0.12 &86.67$\pm$0.38 & \textbf{93.89}$\pm$0.34 & \textbf{91.86}$\pm$0.25\\
		\specialrule{.1em}{.05em}{.05em}
	\end{tabular}
	\caption{Comparison between \textbf{dynamic residual transfer (DRT)} with the state-of-the-art models on Digit-five dataset. The source domains and target domain are shown at the top of each column. }
	\label{tab:digit5}
	\vspace{-1.0em}
\end{table*}
\noindent{\textbf{Evaluation on Digit Five Dataset:}} Table \ref{tab:digit5} shows a comparison to $6$ baselines on Digit Five. DRT outperforms all other methods, beating the state of the art (CMSS) by more than $1\%$, despite a much simpler implementation. Comparing performance in individual adaptation problems, DRT has the best performance on four of the five problems considered. The only exception occurs when SVHN is the target domain, where DRT achieves the second best performance of all methods. Beyond this, the smallest gains occur when MNIST is the target domain. This was expected, since MNIST is easier to transfer to and somewhat saturated. In general, the gains of DRT increase with domain discrepancy, reaching $5.7\%$ for the hardest transfer problem (MNIST-M as target domain).

\begin{table*}[t]
	\centering
	\small
	\setlength{\tabcolsep}{2.5pt}
	\begin{tabular}{cccccccc}
		\specialrule{.1em}{.05em}{.05em}
		Models & \begin{tabular}{cc}inf,pnt,qdr \\ rel,skt $\rightarrow$ clp\end{tabular}& \begin{tabular}{cc}clp,pnt,qdr \\ rel,skt $\rightarrow$ inf\end{tabular} & \begin{tabular}{cc}clp,inf,qdr \\ rel,skt $\rightarrow$ pnt\end{tabular} & \begin{tabular}{cc}clp,inf,pnt \\ rel,skt $\rightarrow$ qdr\end{tabular} & \begin{tabular}{cc}clp,inf,pnt \\ qdr,skt $\rightarrow$ rel\end{tabular} & \begin{tabular}{cc}clp,inf,pnt \\ qdr,rel $\rightarrow$ skt\end{tabular} & Avg\\
		\hline
		Source Only & 52.1$\pm$0.51 & 23.4$\pm$0.28 & 47.7$\pm$0.96 & 13.0$\pm$0.72 & 60.7$\pm$0.23 & 46.5$\pm$0.56 & 40.6$\pm$0.56\\
		\hline
		ADDA \cite{tzeng2017adversarial} & 47.5$\pm$0.76 & 11.4$\pm$0.67 & 36.7$\pm$0.53 & 14.7$\pm$0.50 & 49.1$\pm$0.82 & 33.5$\pm$0.49 & 32.2$\pm$0.63\\
		MCD \cite{saito2018maximum} & 54.3$\pm$0.64& 22.1$\pm$0.70 & 45.7$\pm$0.63 & 7.6$\pm$0.49 & 58.4$\pm$0.65 & 43.5$\pm$0.57 & 38.5$\pm$0.61\\
		DANN \cite{ganin2015unsupervised} & 60.6$\pm$0.42 & 25.8$\pm$0.43 & 50.4$\pm$0.51 & 7.7$\pm$0.68 & 62.0$\pm$0.66 & 51.7$\pm$0.19 & 43.0$\pm$0.46\\
		\hline
		DCTN \cite{xu2018deep} & 48.6$\pm$0.73 & 23.5$\pm$0.59 & 48.8$\pm$0.63 & 7.2$\pm$0.46 & 53.5$\pm$0.56 & 47.3$\pm$0.47 & 38.2$\pm$0.57\\
		M$^3$SDA-$\beta$ \cite{peng2019moment} & 58.6$\pm$0.53 & 26.0$\pm$0.89 & 52.3$\pm$0.55 & 6.3$\pm$0.58 & 62.7$\pm$0.51 & 49.5$\pm$0.76 & 42.6$\pm$0.64\\
        ML-MSDA \cite{li2020mutual} & 61.4$\pm$0.79 & 26.2$\pm$0.41 & 51.9$\pm$0.20 & \textbf{19.1}$\pm$0.31 & 57.0$\pm$1.04 & 50.3$\pm$0.67 & 44.3$\pm$0.24 \\
        Meta-MCD \cite{li2020online} & 62.8$\pm$0.22 & 21.4$\pm$0.07 & 50.5$\pm$0.08 & 15.5$\pm$0.22 & 64.6$\pm$0.16 & 50.4$\pm$0.12 & 44.2$\pm$0.07 \\
        LtC-MSDA  \cite{wang2020learning} & 63.1$\pm$0.5 & 28.7$\pm$0.7 & 56.1$\pm$0.5 & 16.3$\pm$0.5 & 66.1$\pm$0.6 & 53.8$\pm$0.6 & 47.4$\pm$0.6 \\
		CMSS \cite{yang2020curriculum} & 64.2$\pm$0.18 & 28.0$\pm$0.20 & 53.6$\pm$0.39 & 16.0$\pm$0.12 & 63.4$\pm$0.21 & 53.8$\pm$0.35 & 46.5$\pm$0.24 \\
		\hline
		\textbf{DRT} & 69.7$\pm$0.24& 31.0$\pm$0.56 & 59.5$\pm$0.43 & 9.9$\pm$1.03 & 68.4$\pm$0.28 & 59.4$\pm$0.21 & 49.7$\pm$0.46\\
		\textbf{DRT+ST} & \textbf{71.0}$\pm$0.21 & \textbf{31.6}$\pm$0.44 & \textbf{61.0}$\pm$0.32 & 12.3$\pm$0.38 & \textbf{71.4}$\pm$0.23 & \textbf{60.7}$\pm$0.31 & \textbf{51.3}$\pm$0.32\\
		\specialrule{.1em}{.05em}{.05em}
	\end{tabular}
	\caption{Comparison between \textbf{dynamic residual transfer (DRT)} with the state-of-the-art models on DomainNet. (`DRT+ST' represents the combination between dynamic residual transfer and self-training for domain adaptation)}
	\label{tab:domainnet}
\end{table*}

\noindent{\textbf{Evaluation on DomainNet Dataset. }} For DomainNet \cite{peng2019moment}, a ResNet-101 \cite{he2016deep} was used as backbone and DRT was compared to $11$ baselines. Among these, ADDA \cite{tzeng2017adversarial}, DANN \cite{ganin2015unsupervised} and  MCD \cite{saito2018maximum} were developed for traditional unsupervised domain adaptation (UDA), where a single-source domain is assumed. The remaining are multi-source domain adaptation methods that require domain labels.
 
Table \ref{tab:domainnet} shows that DRT improves on the state-of-the-art method- CMSS by more than $3\%$ (49.7\% vs 46.5\%). When DRT is combined with a naive self-training method (DRT+ST), it achieves gains of $3.9\%$ over LtC-MSDA \cite{wang2020learning}, a methods that generates pseudo-labels for the target samples  (during self-training, pseudo-labels for target samples with confidence greater than $0.8$ are used to train DRT again with source samples).
Compared to the adaptation methods that use no domain labels (single-source), DRT improves the best average adaptation results (DANN) by $6.7\%$ (49.7\% vs. 43\%).

Regarding individual adaptation problems, DRT achieves the best performance for all target domains other than `quickdraw'. This can be explained by the large domain gap between `quickdraw' and the other domains, and the fact that the MCD loss does not fare well in this problem. Better results would likely be possible by using the M$^3$SDA loss, as was the case in Table \ref{tab:loss_abl}.

\begin{table*}[t]
	\centering
	\small
	\setlength{\tabcolsep}{3.5pt}
	\begin{tabular}{cccccccc}
	\specialrule{.1em}{.05em}{.05em}
		Models & \begin{tabular}{cc}inf,pnt,qdr \\ rel,skt $\rightarrow$ clp\end{tabular}& \begin{tabular}{cc}clp,pnt,qdr \\ rel,skt $\rightarrow$ inf\end{tabular} & \begin{tabular}{cc}clp,inf,qdr \\ rel,skt $\rightarrow$ pnt\end{tabular} & \begin{tabular}{cc}clp,inf,pnt \\ rel,skt $\rightarrow$ qdr\end{tabular} & \begin{tabular}{cc}clp,inf,pnt \\ qdr,skt $\rightarrow$ rel\end{tabular} & \begin{tabular}{cc}clp,inf,pnt \\ qdr,rel $\rightarrow$ skt\end{tabular} & Avg\\
		\hline
		ADDA \cite{tzeng2017adversarial} & 28.2/39.5 & 9.3/14.5 & 20.1/29.1 & {\bf{8.4/14.9}} & 31.1/41.9 & 21.7/30.7 & 19.8/28.4\\
		\hline
		MCD \cite{saito2018maximum} & 31.4/42.6 & 13.1/19.6 & 24.9/42.6 & 2.2/3.8 & 35.7/50.5 & 23.9/33.8 & 21.9/32.2\\
		\hline
		\textbf{DRT} & {\bf{41.9/56.2}} & {\bf{19.6/26.6}} & {\bf{35.3/53.4}} & 8.0/12.2 & {\bf{44.5/55.5}} & {\bf{35.0/44.8}} & {\bf{30.7/41.5}}\\
		\hline
	\end{tabular}
	\caption{\textbf{Single source domain adaptation performance on DomainNet}. Each column, shows the average/best classification accuracy for transfer from all source to the specified target domain.}
	\label{tab:avg_best_single}
\end{table*}

\subsection{Single-Source to Single-Target Adaptation}
The performance of DRT on the traditional domain adaptation problem (single-source domain) was also evaluated on DomainNet \cite{peng2019moment}. In this case, for each target domain, adaptation was performed from each of the other five domains (sources). The average and best performance among these adaptations is shown in Table~\ref{tab:avg_best_single}, for each target domain.
DRT again significantly outperforms the previous domain adaptation methods. For example, when `clipart' is the target domain, its average adaptation performance is $10.5\%$ better than that of the MCD method. On average, over all pairs of source and target domains, it outperforms MCD by more than $8\%$. These results show that, for problems with hundreds of classes, dynamic residual transfer can lead to very large adaptation gains even in the traditional domain adaptation setting. This confirms the claim that even these problems tend to have many sub-domains. When this is the case, the ability of dynamic residual transfer to adapt the model on a per-example basis can be a significant asset.

Finally, comparing the results of Tables \ref{tab:domainnet} and \ref{tab:avg_best_single} shows that DRT trained on multi-source domains performs $8.2\%$ better ($49.7\%$ vs. $41.5\%$) than the average of the best single-source domain transfers. This improvement is about $2\%$ better than that given by MCD ($8.2\%$ vs. $6.3\%$). This shows that considering a variety of source domains improves domain adaptation performance, especially when dynamic residual transfer is used. A main advantage of DRT is that it can be applied to all settings, since it does not require domain labels. Its universal nature makes it irrelevant if the problem is single-source or multi-source. It suffices to collect training data and DRT will automatically figure out how to adapt the network to all settings. There is no need to even define ``source domains."

\begin{figure}[t]
\begin{subfigure}{0.23\textwidth}
   \centering
   \includegraphics[width=0.98\linewidth]{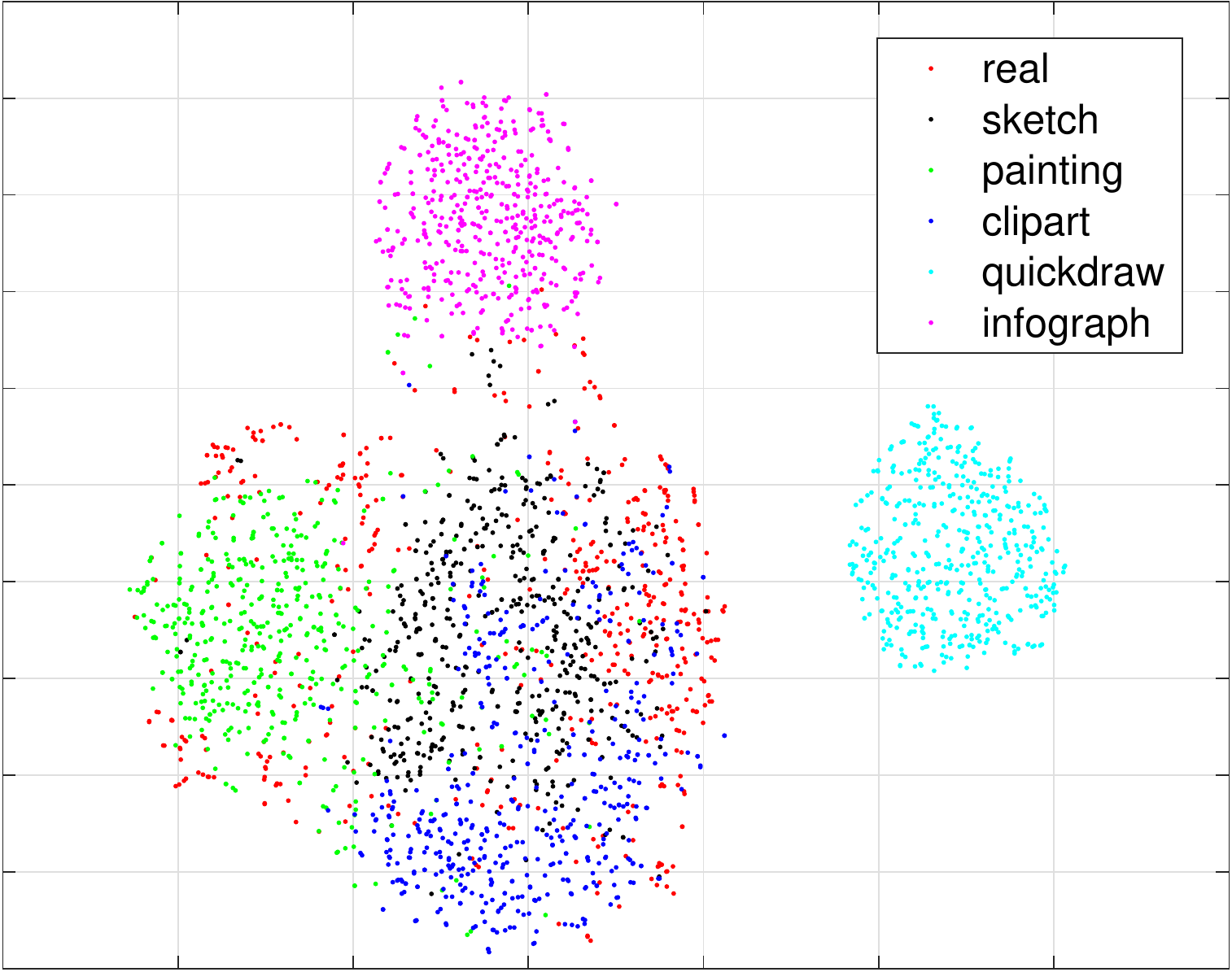}
   \caption{target: clipart}
\end{subfigure}
\begin{subfigure}{0.23\textwidth}
   \centering
   \includegraphics[width=0.98\linewidth]{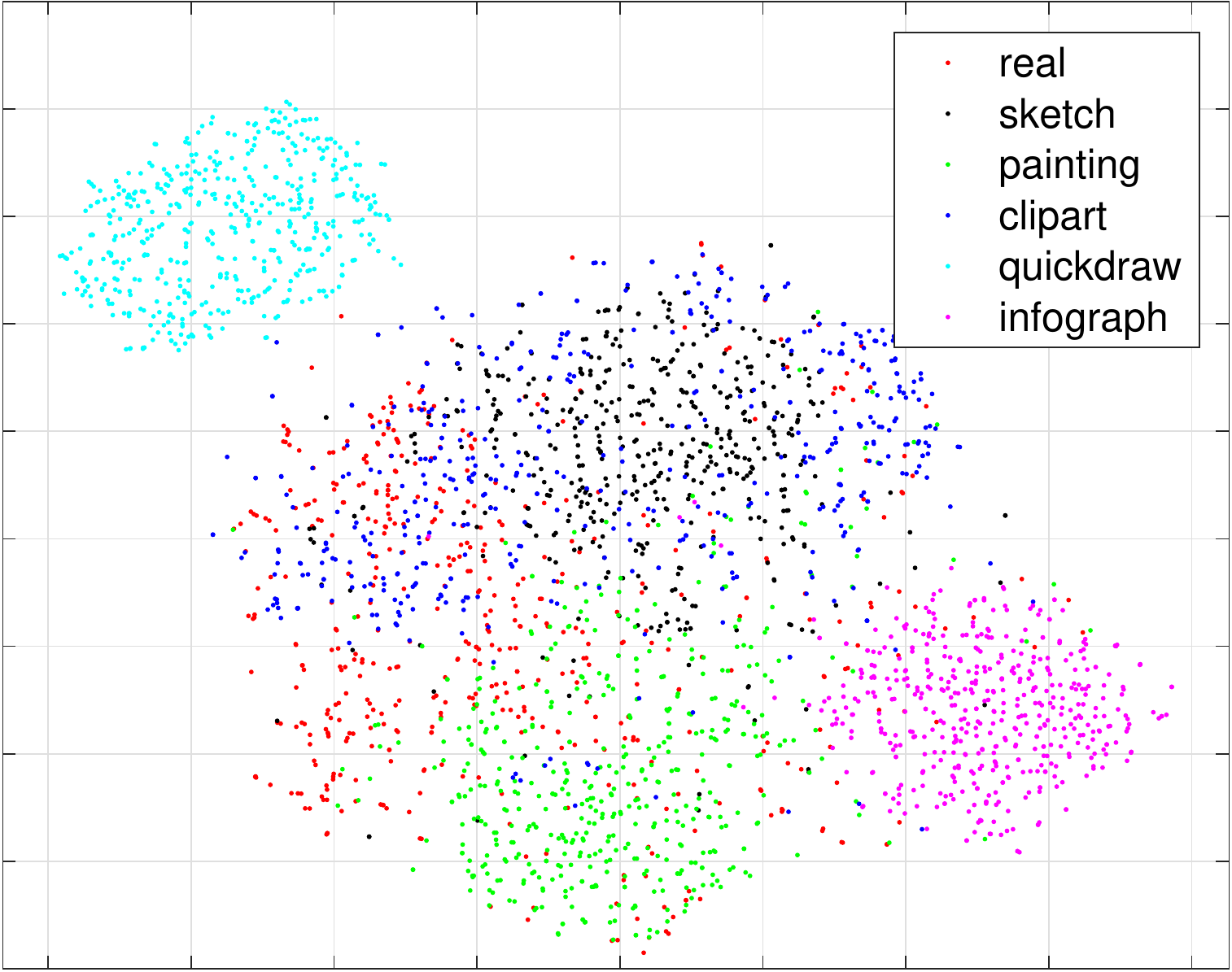}
   \caption{target: infograph}
\end{subfigure}

\begin{subfigure}{0.23\textwidth}
   \centering
   \includegraphics[width=0.98\linewidth]{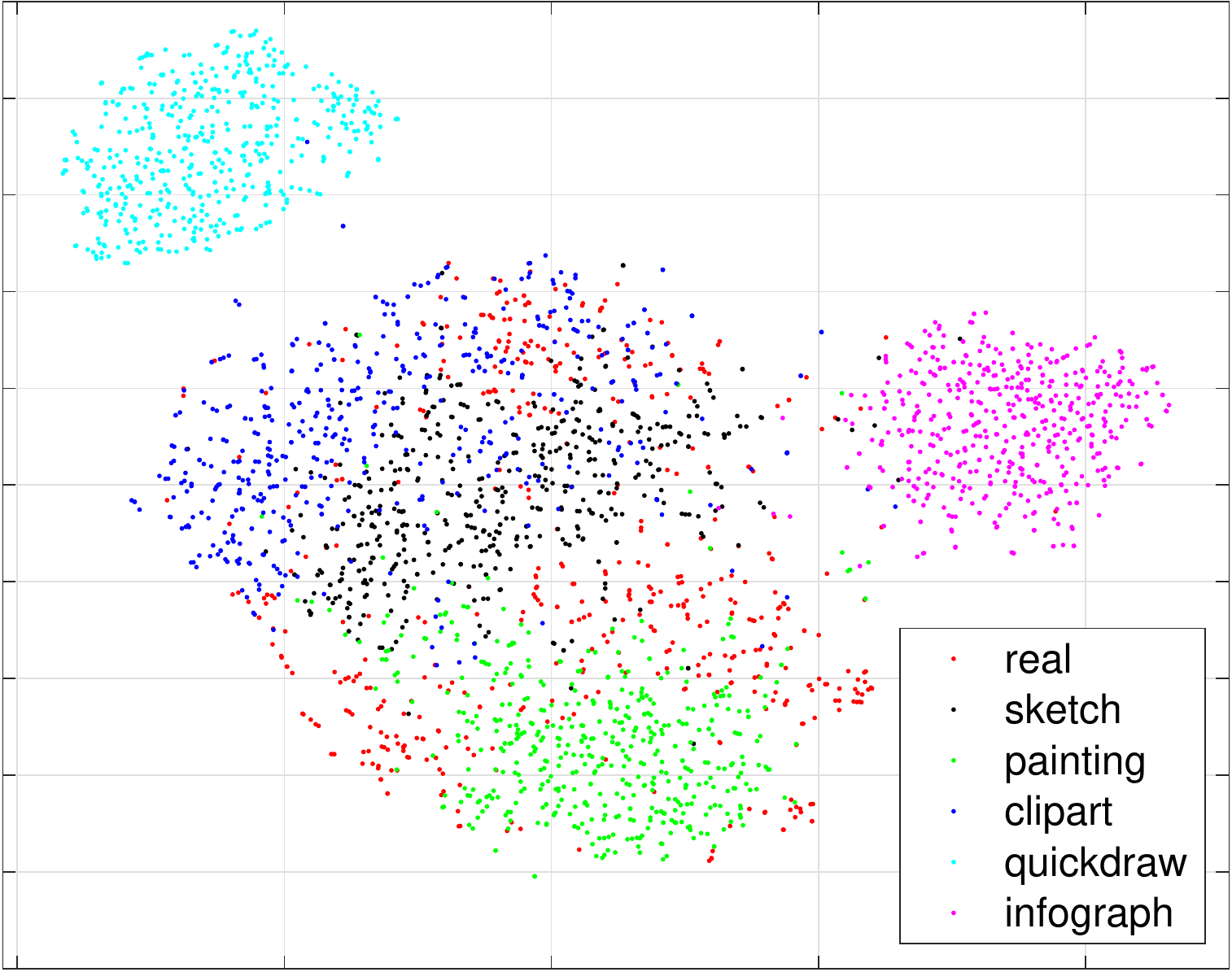}
   \caption{target: painting}
\end{subfigure}
\begin{subfigure}{0.23\textwidth}
   \centering
   \includegraphics[width=0.98\linewidth]{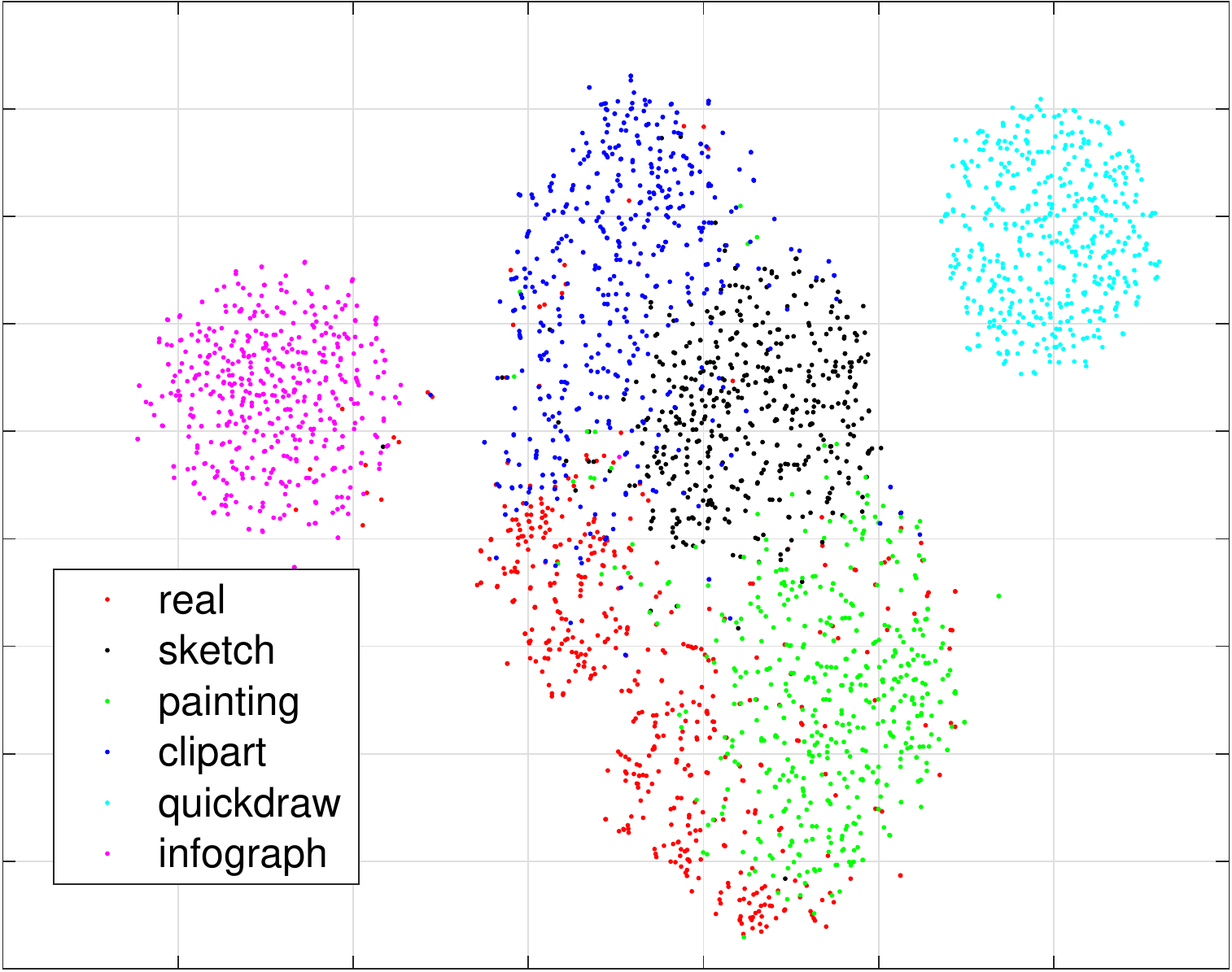}
   \caption{target: quickdraw}
\end{subfigure}

\begin{subfigure}{0.23\textwidth}
   \centering
   \includegraphics[width=0.98\linewidth]{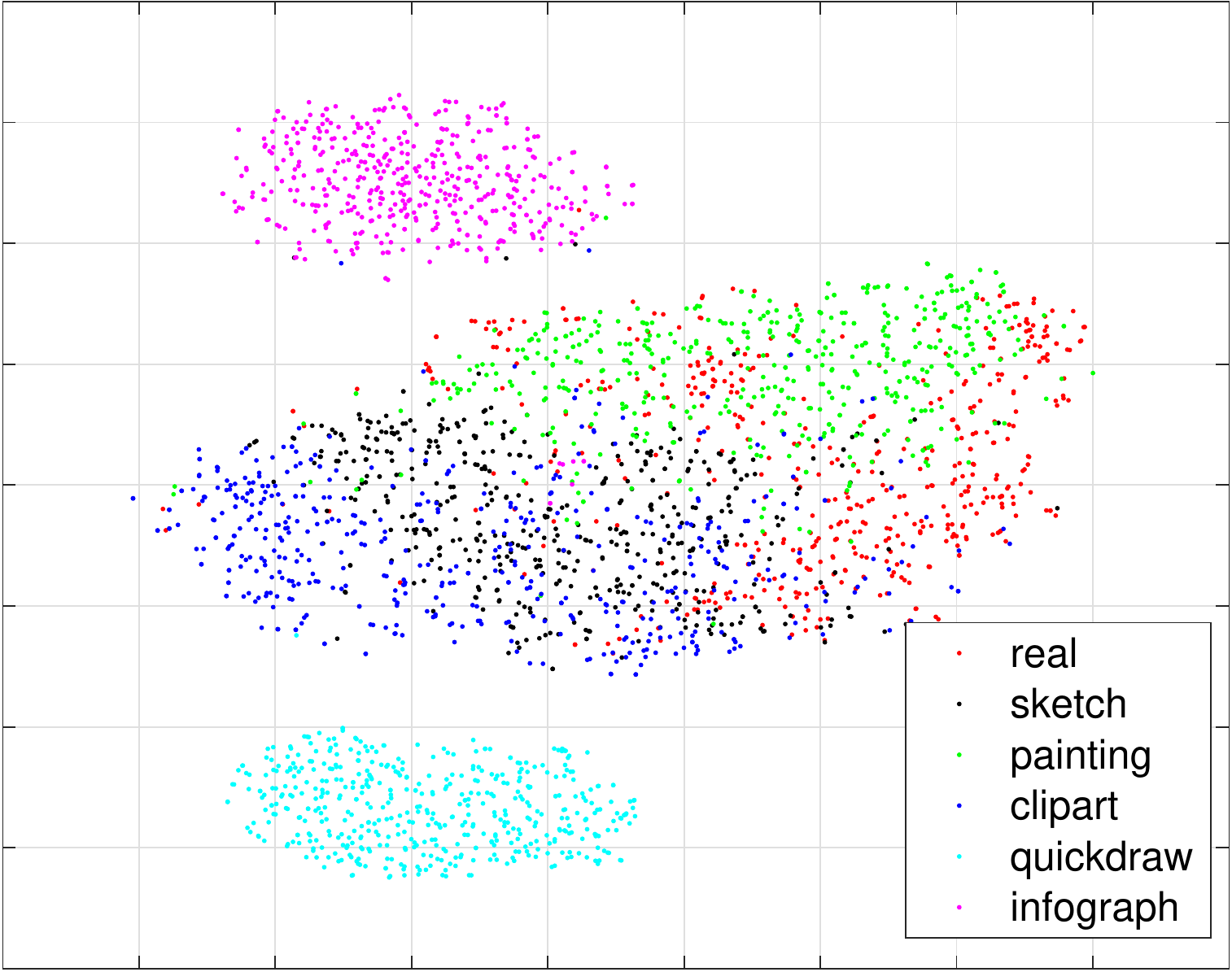}
   \caption{target: real}
\end{subfigure}
\begin{subfigure}{0.23\textwidth}
   \centering
   \includegraphics[width=0.98\linewidth]{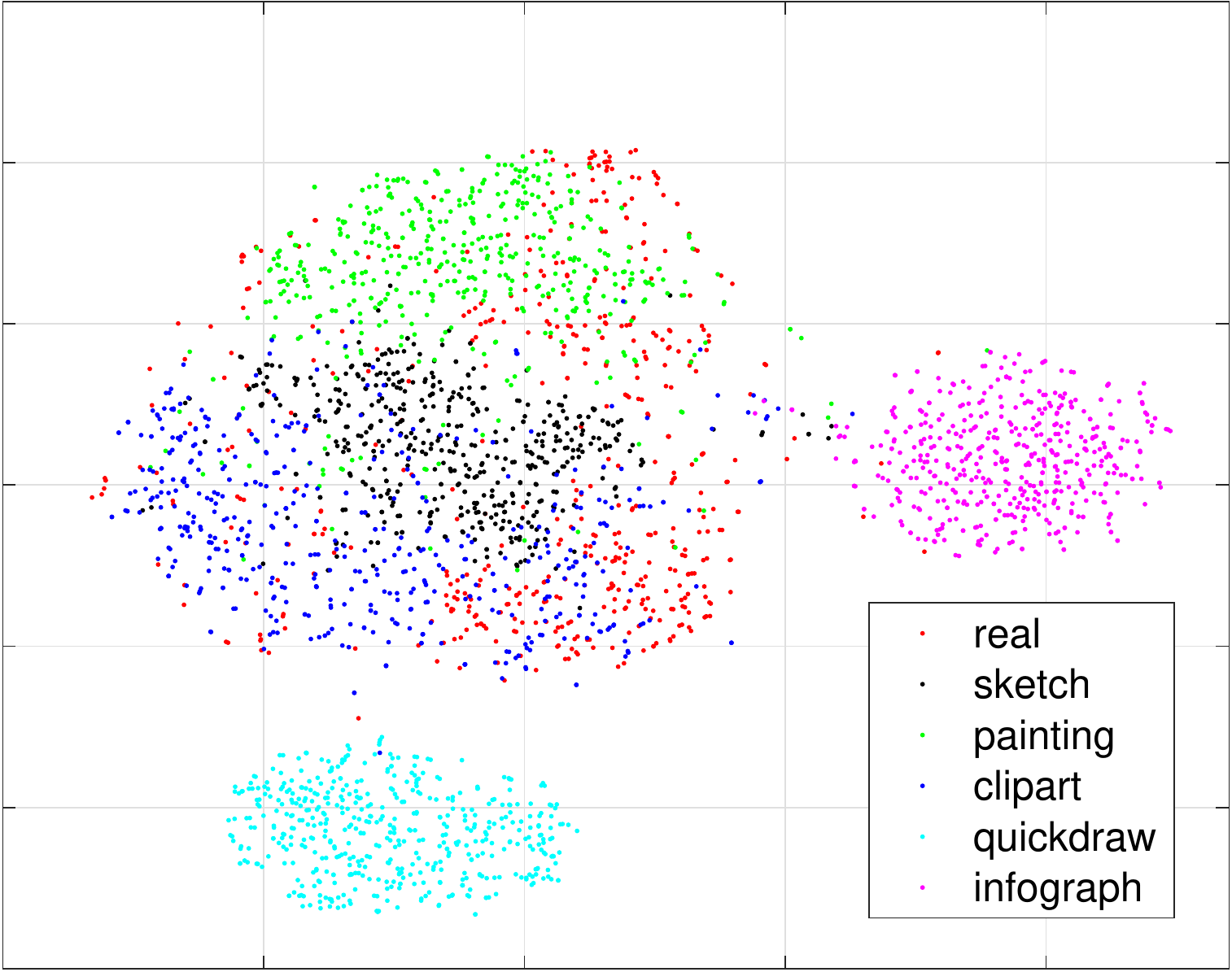}
   \caption{target: sketch}
\end{subfigure}
\caption{The t-SNE \textbf{visualization of dynamic coefficients} $\Pi=\{\pi_i^l(\bm{x})\}$ when DRT is trained with target domain- `clipart', `infograph', `painting', `quickdraw', `real' and `sketch'. (Best view in color)}
\label{fig:tsne}
\vspace{-2.0em}
\end{figure}

\begin{figure}[t]
\vspace{0.2em}
\begin{subfigure}{0.23\textwidth}
   \centering
   \includegraphics[width=0.98\linewidth]{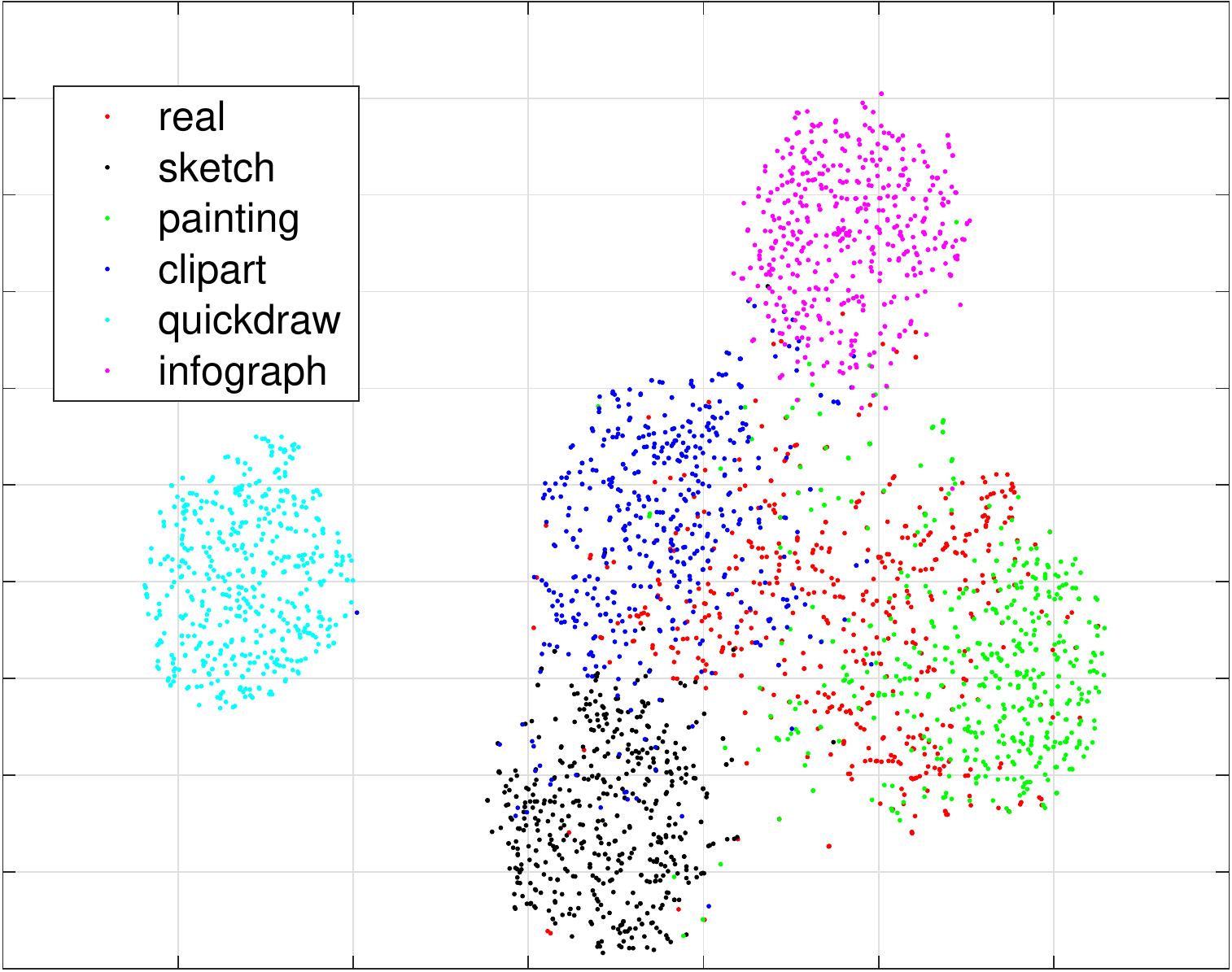}
   \caption{target: clipart ($\Pi_{low}$)}
\end{subfigure}
\begin{subfigure}{0.23\textwidth}
   \centering
   \includegraphics[width=0.98\linewidth]{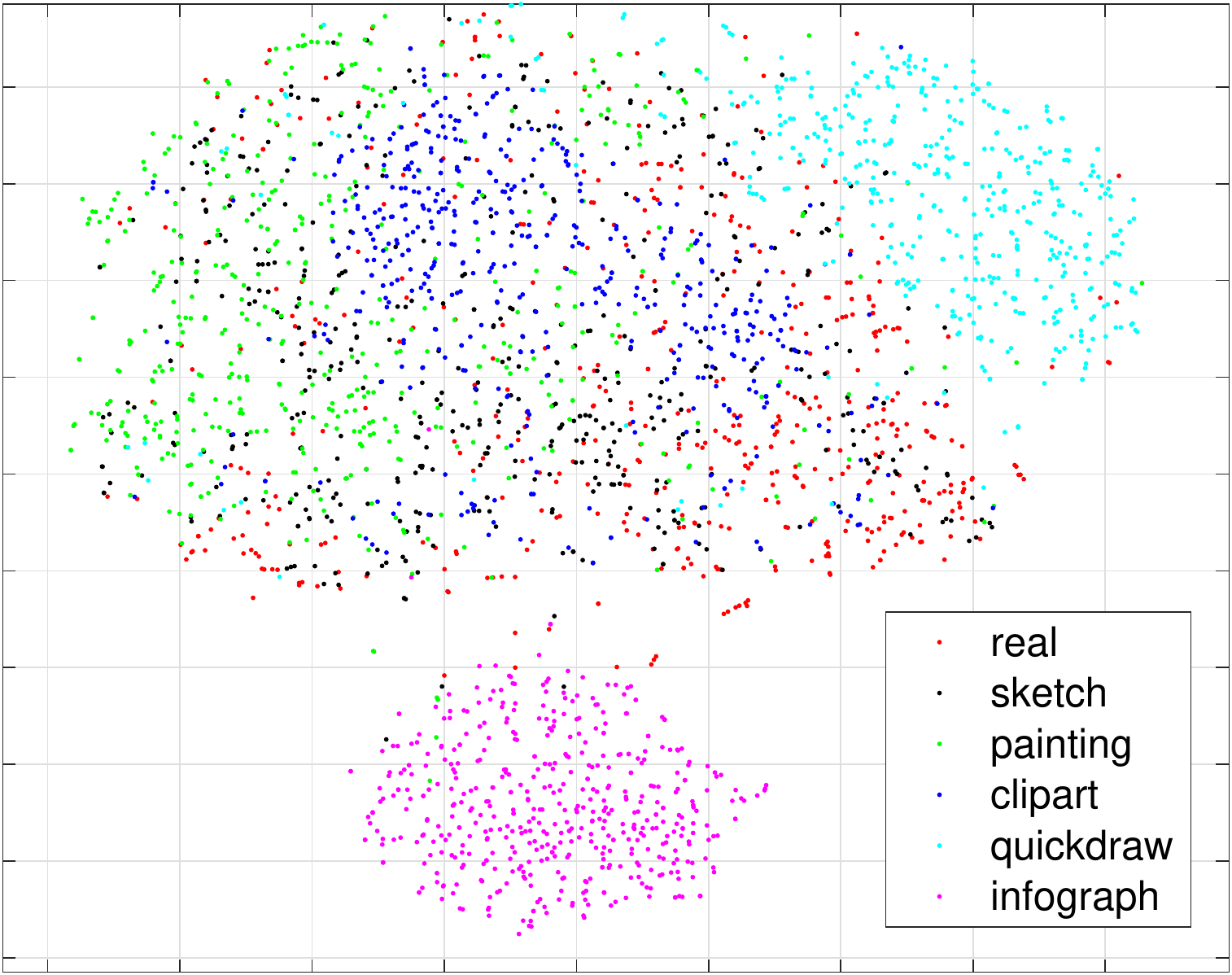}
   \caption{target: clipart ($\Pi_{high}$)}
\end{subfigure}

\begin{subfigure}{0.23\textwidth}
   \centering
   \includegraphics[width=0.98\linewidth]{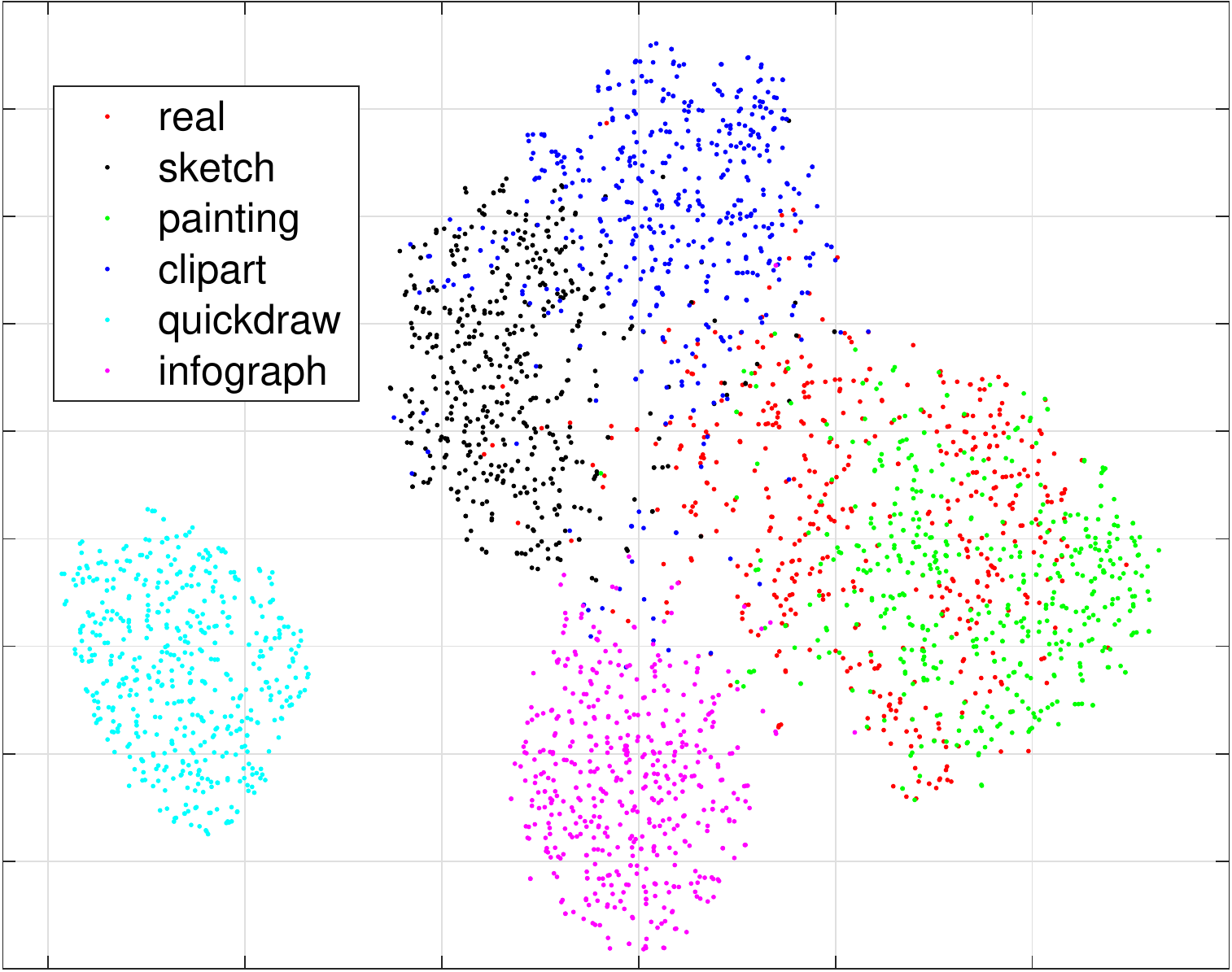}
   \caption{target: real ($\Pi_{low}$)}
\end{subfigure}
\begin{subfigure}{0.23\textwidth}
   \centering
   \includegraphics[width=0.98\linewidth]{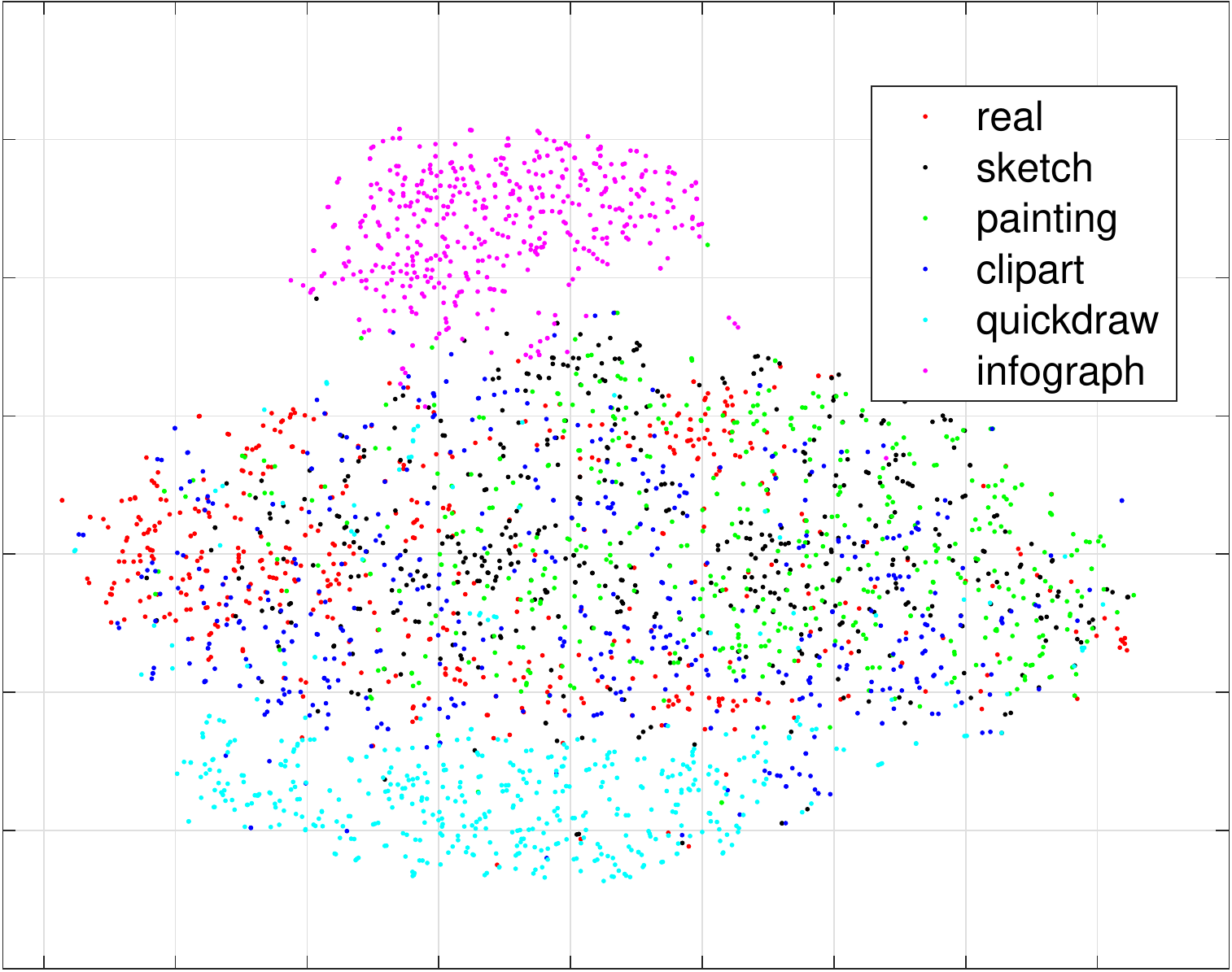}
   \caption{target: real ($\Pi_{high}$)}
\end{subfigure}
\caption{The t-SNE visualization of \textbf{first half} ($\Pi_{low}$) and \textbf{second half} ($\Pi_{high}$) dynamic coefficients when DRT is trained with target domain- `clpart' and `real'. (Best view in color)}
\label{fig:tsne_half}
\vspace{-2.0em}
\end{figure}

\subsection{Visualization}

To obtain further insight about dynamic residual transfer (DRT), we visualized the dynamic coefficients of Equation \ref{equ:dy_res} with t-SNE \cite{maaten2008visualizing}. For each sample, we created a vector $\Pi=\{\pi_i^l(\bm{x})\}, i\in\{1,2,...K\}, l\in\{1,2,...L\}$ by concatenating the dynamic coefficients from the $L$ network layers. The vectors $\Pi$ from different target domains are visualized in Figure \ref{fig:tsne}. A more detailed visualization is given in Figure \ref{fig:tsne_half}, by splitting $\Pi$ into $\Pi_{low}$ and $\Pi_{high}$, which include the coefficients from lower and higher network layers. For brevity, Figure \ref{fig:tsne_half}, only visualizes the model trained with `clipart' and `real' as target domain and the other domains show similar trend. 

Figure \ref{fig:tsne} first shows that domain information is embedded into the dynamic coefficients $\{\pi_i^l(\bm{x})\}$. This can be observed by samples from same domains tend to group in identifiable clusters, which confirms our claim that adapting model across domains can be achieved through adapting model to samples. 
Secondly, the distance among clusters reflects domain shifts that explain how adaptation performance varies with target domain. For example, the fact that the dynamic coefficients of `quickdraw' are always quite different from others, explains why adaptation performance is weaker when this is the target domain. Thus for `quickdraw', either a more powerful alignment loss is needed to shift the samples close enough to the source domains to enable the dynamic model adapted to this domain or a more complex dynamic model e.g. combination of `channel attention' and `subspace routing' is required.
Figure \ref{fig:tsne_half} further shows that the dynamic coefficients from lower layers (Figure \ref{fig:tsne_half}(a)) form much more clear domain clusters than the coefficients of the higher layers (Figure \ref{fig:tsne_half}(b)). This shows that the network features become more domain agnostic in the higher layers, confirming the effectiveness of DRT to reduce domain discrepancies.   

\section{Conclusion}
In this paper, we introduce dynamic transfer for multi-source domain adaptation, in which the model parameters are not static but adaptive to input samples. Dynamic transfer mitigates conflicts across multiple domains and unifies multiple source domains into a single source domain, which simplifies the alignment between source and target domains. Experimental results show that dynamic transfer achieves a better adaptation performance compared to the state-of-the-art method for multi-source domain adaptation. We hope this paper can give a new understanding about multi-source domain adaptation.

{\small
\bibliographystyle{ieee_fullname}
\bibliography{egbib}
}

\end{document}